\pdfoutput=1
\documentclass[11pt]{article}

\usepackage[final]{acl}

\usepackage{amsmath}
\usepackage{times}
\usepackage{booktabs}
\usepackage{graphicx}
\usepackage{threeparttable}
\usepackage{latexsym}
\usepackage[T1]{fontenc}
\usepackage[utf8]{inputenc}
\usepackage{microtype}
\usepackage{inconsolata}
\usepackage{array}
\usepackage{graphicx}
\usepackage{subcaption}
\usepackage{listings}
\usepackage{tcolorbox}
\usepackage{hwemoji}
\colorlet{punct}{red!60!black}
\definecolor{background}{HTML}{EEEEEE}
\definecolor{delim}{RGB}{20,105,176}
\colorlet{numb}{magenta!60!black}
\lstdefinelanguage{json}{
    basicstyle=\normalfont\ttfamily,
    numbersep=4pt,
    breaklines=true,
    backgroundcolor=\color{background},
    literate=
     *{0}{{{\color{numb}0}}}{1}
      {1}{{{\color{numb}1}}}{1}
      {2}{{{\color{numb}2}}}{1}
      {3}{{{\color{numb}3}}}{1}
      {4}{{{\color{numb}4}}}{1}
      {5}{{{\color{numb}5}}}{1}
      {6}{{{\color{numb}6}}}{1}
      {7}{{{\color{numb}7}}}{1}
      {8}{{{\color{numb}8}}}{1}
      {9}{{{\color{numb}9}}}{1}
      {:}{{{\color{punct}{:}}}}{1}
      {,}{{{\color{punct}{,}}}}{1}
      {\{}{{{\color{delim}{\{}}}}{1}
      {\}}{{{\color{delim}{\}}}}}{1}
      {[}{{{\color{delim}{[}}}}{1}
      {]}{{{\color{delim}{]}}}}{1},
}

\usepackage{siunitx} 
\usepackage{graphicx}
\usepackage{textcomp}
\usepackage{color,soul}
\usepackage{bm}
\usepackage{xcolor}
\usepackage{colortbl}
\usepackage{multirow}

\usepackage{tcolorbox}
\tcbuselibrary{breakable}
\usepackage{enumitem}

\definecolor{Community}{HTML}{d8b365}
\definecolor{NormVio}{HTML}{5ab4ac}
\definecolor{ForestGreen}{RGB}{0,77,36}

\colorlet{tableheadcolor}{gray!25}

\colorlet{tablerowcolor}{gray!15}
\colorlet{tablerowcolor2}{gray!12}
\colorlet{tablerowcolor3}{gray!25}
\colorlet{tablerowcolor4}{gray!50}

\definecolor{lred}{HTML}{fbb4ae}
\definecolor{lblue}{HTML}{b3cde3}
\definecolor{lgreen}{HTML}{ccebc5}
\definecolor{lviolet}{HTML}{decbe4}
\definecolor{lorange}{HTML}{fed9a6}
\definecolor{lyellow}{HTML}{ffffcc}

\newcommand{\allocate}{\colorbox{pink!50}{\textbf{\texttt{Allocate}}}}
\newcommand{\predict}{\colorbox{orange!40}{\textbf{\texttt{Predict}}}}
\newcommand{\aggregate}{\colorbox{cyan!25}{\textbf{\texttt{Aggregate}}}}
\newcommand{\explain}{\colorbox{green!30}{\textbf{\texttt{Explain}}}}

\newcommand{\momo}{\texttt{MoMoE}}
\newcommand{\momoco}{\texttt{MoMoE\textsubscript{Community}}}
\newcommand{\momonv}{\texttt{MoMoE\textsubscript{NormVio}}}

\title{MoMoE: \underline{M}ixture \underline{o}f \underline{Mo}deration \underline{E}xperts Framework\\ for AI-Assisted Online Governance}

\author{
Agam Goyal, 
Xianyang Zhan\textsuperscript{$\ast$},
Yilun Chen\textsuperscript{$\ast$},
Koustuv Saha\textsuperscript{\textdaggerdbl},
Eshwar Chandrasekharan\textsuperscript{\textdaggerdbl} \\
Siebel School of Computing and Data Science\\
University of Illinois Urbana-Champaign\\
 \texttt{\{agamg2, zhan39, yilunc3, ksaha2, eshwar\}@illinois.edu}
}

\begin{document}
\maketitle
\def\thefootnote{$\ast$}\footnotetext{Both authors contributed equally.}\def\thefootnote{\arabic{footnote}}
\def\thefootnote{\textdaggerdbl}\footnotetext{Both authors are advisors of this work.}\def\thefootnote{\arabic{footnote}}

\begin{abstract}
Large language models (LLMs) have shown great potential in flagging harmful content in online communities. Yet, existing approaches for moderation require a separate model for every community and are opaque in their decision-making, limiting real-world adoption. We introduce \textbf{\underline{M}}ixture \textbf{\underline{o}}f \textbf{\underline{Mo}}deration \textbf{\underline{E}}xperts (\momo{}), a modular, cross-community framework that adds post-hoc explanations to scalable content moderation. \momo{} orchestrates four operators---\allocate{}, \predict{}, \aggregate{}, \explain{}---and is instantiated as seven community-specialized experts (\momoco{}) and five norm-violation experts (\momonv{}). On 30 unseen subreddits, the best variants obtain Micro-F1 scores of 0.72 and 0.67, respectively, matching or surpassing strong fine-tuned baselines while consistently producing concise and reliable explanations. Although community-specialized experts deliver the highest peak accuracy, norm-violation experts provide steadier performance across domains. These findings show that \momo{} yields scalable, transparent moderation without needing per-community fine-tuning. More broadly, they suggest that lightweight, explainable expert ensembles can guide future NLP and HCI research on trustworthy human-AI governance of online communities.\footnote{Code: \href{https://github.com/scuba-illinois/MoMoE}{https://github.com/scuba-illinois/MoMoE}}
\end{abstract}

\section{Introduction}

A persistent challenge that online communities face is identifying content that violates community norms. This challenge is particularly crucial on platforms like Reddit, which hosts over 125,000 active communities called subreddits with diverse norms and moderation needs, placing significant burden on unpaid moderators~\cite{li_measuring_2022}. 

To alleviate this burden, various sociotechnical tools for content moderation have been proposed in prior work. These include keyword-based moderation using simple regular expression filters~\cite{long2017could,jhaver_human-machine_2019,jhaver_designing_2022}, traditional ML-based moderation, which range from embedding-based classifiers~\cite{chandrasekharan_bag_2017, chandrasekharan_crossmod_2019} to language model (LM)-based moderation approaches, which have recently gained popularity as they show promising performance and can enhance transparency.~\cite{kumar2024watch, 10.1145/3613905.3650828,zhan-etal-2025-slm}.

However, existing LM-based approaches for content moderation face some key challenges that hinder their deployment in real-world scenarios.

First, while \citet{zhan-etal-2025-slm} demonstrated that fine-tuned SLMs can outperform off-the-shelf LLMs on content moderation, they require substantial community-specific training data for fine-tuning models. This creates significant barriers for new communities, as they may lack historical moderation data required to fine-tune these models. 

Second, research has identified that different online communities operate under shared yet distinct norms and values~\cite{chandrasekharan_internets_2018, goyal_uncovering_2024}. Yet, existing LM-based approaches rely on instantiating a single model per community, which hinders the ability of these models to cater to a large number of communities that may share a similar kind of norms violations, to enable a cross-community moderation approach.

Third, while existing approaches focus solely on accuracy, recent work has called for improved transparency in order to improve moderator trust~\cite{huang2024content,palla2025policy,moran2025end} and ensure that moderator workload doesn't increase due to difficulty in identifying inconsistencies within these systems~\cite{ashkinaze2024seeing}.

Finally, current AI-based approaches treat content moderation as a fully automated task, overlooking the crucial role of human moderators who possess contextual understanding and community expertise. Effective moderation systems should not aim to replace human moderators but rather augment and complement their capabilities by providing 
transparent justifications that allow for human oversight and intervention~\cite{10.1145/3287560.3287598,10.1145/3613905.3650828,10.1145/3643834.3661623}. 
Therefore, there is a critical need for frameworks that can show how to efficiently operate with these AI-based approaches by leveraging the complementary strengths of humans and LLMs. 

In this paper, we introduce \momo{} (\textbf{\underline{M}}ixture \textbf{\underline{o}}f \textbf{\underline{Mo}}deration \textbf{\underline{E}}xperts), a novel ensemble framework for cross-community content moderation that addresses these limitations. \momo{} is a modular framework that is composed of four operators: 
\begin{figure*}[t]
    \centering
    \includegraphics[width=\linewidth]{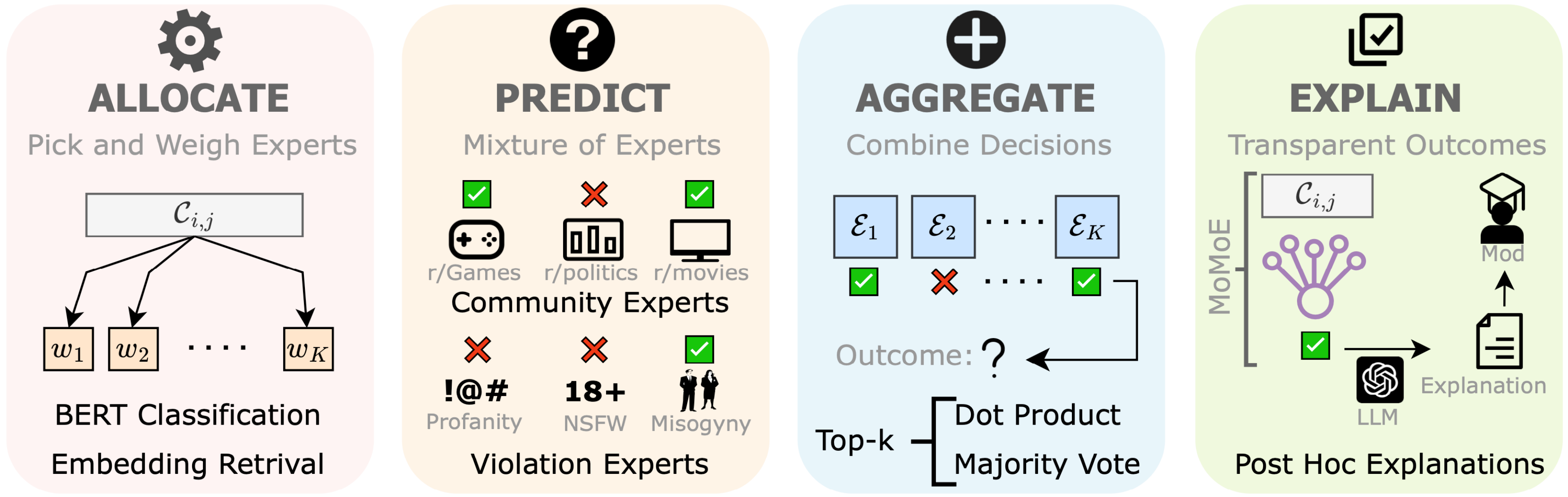}
    \caption{\textbf{\momo{}} is composed of four modular operators--- \textbf{(1)\allocate{}:} Determines how to pick the relevant experts and weigh the predictions they provide using softmax probabilities from classification models or similarity-based scoring; \textbf{(2)\predict{}:} Determines individual expert predictions from two kind of ensembles, with community-specific experts or norm-violation experts; \textbf{(3)\aggregate{}:} Determines how to aggregate the predictions of individual experts into a single outcome using strategies like dot product between allocated weights and expert predictions or majority voting; and \textbf{(4)\explain{}:} Uses a post hoc LLM-based approach to summarize and explain \momo{}'s decision output to help moderators understand outcomes and rectify potential inconsistencies.}
    \label{fig:momoe-framework}
\end{figure*}
\noindent\textbf{(1) \allocate{}:} Operator that decides how to pick relevant experts and weigh their decisions for a specific instance of the moderation task (e.g., classification-based, similarity-based, etc.); \noindent\textbf{(2) \predict{}:} Operator that leverages a mixture of fine-tuned small language models (``experts'') representing either community-based experts (\momoco{}) or norm violation-based experts (\momonv{}) to provide a moderation outcome; \noindent\textbf{(3) \aggregate{}:} Operator that decides how to combine predictions of the individual experts (e.g., dot-product composition, majority voting, etc.); and \noindent\textbf{(4) \explain{}:} Operator that provides simplified post-hoc LLM-based explanations for \momo{} decisions.

We evaluate the effectiveness of \momo{} using a comment removal dataset~\cite{chandrasekharan_hybrid_2019} by simulating a real-time content moderation scenario, and perform an extensive quantitative and qualitative analysis. 
We find that \momo{} performs competitively against strong baselines on $30$ unseen communities. Specifically, the best configurations of \momoco{} and \momonv{} achieve Micro-F1 scores of $0.72$ and $0.67$, respectively. While \momoco{} achieves a wider range of performance depending on the target community, \momonv{} provides consistently strong performance across communities. Through case studies, we provide a detailed analysis of the complementary strengths of different operator configurations. Further, through manual inspection we find that the explanations provided by \explain{} reliably reflect the decision-making trace of \momo{}.

By integrating multiple expert perspectives and providing transparent explanations, \momo{} aims to create a more generalizable approach to AI-assisted governance of online forums that upholds community-specific norms while leveraging cross-community knowledge. The modular nature of \momo{} provides human moderators the agency to intervene and perform recalibration at the level of each operator, and moreover it also provides opportunity for individual components to be enhanced with advancements in NLP. Our goal is to enhance the potential for human-AI collaborative moderation by contributing a \textit{framework for AI-based tools that complement rather than replace human expertise}, while still performing competitively in comparison to strong baselines.
\section{Related Work}

\paragraph{AI-assisted content moderation:}

Content moderation on most online platforms is primarily done manually by either commercial moderators or unpaid volunteers~\cite{gillespie2018custodians, roberts2019behind,li_measuring_2022}. Prior work has proposed many AI-based approaches to content moderation. This includes both embedding-based classifiers~\cite{chandrasekharan_crossmod_2019,park2021detecting} and LLM-based approaches~\cite{mullick2023content,10.1145/3613905.3650828,kumar2024watch,vishwamitra2024moderating,zhan-etal-2025-slm}. However, it has been found that in highly contextual tasks such as moderation, human judgment is often superior to automated judgment~\cite{jurgens2019just, gorwa2020algorithmic}. Due to the dichotomous nature of this problem, there has been a lack of studies on this front, except that of \citet{park2025llm} which proposes a human-LLM pipeline for cross-cultural hate speech moderation. \textit{Our work is a step in this direction of enhancing AI-assisted moderation. We focus on rule-based content moderation, which encompasses hate-speech moderation but is broader and more reflective of real-world moderation processes.}

\paragraph{Human-AI collaborative decision making:} Human decision-making is highly nuanced and contextual. With the rise of LLMs, there has been an increasing body of research that proposes to use LLMs for high-stakes decision making in domains of healthcare~\cite{benkirane-etal-2025-diagnose}, moderation~\cite{koshy2024venire}, etc. Key considerations in these collaborative tools is to assist human decision-making without replacing them, and the final judgment is that of humans~\cite{steyvers2024three}. As a result, there has recently been a growing body of research~\cite{li2023modeling,vereschak2024trust,li2025text,castaneda2025supporting} building tools and approaches that facilitate these decision-making processes. However, despite the growing interest in using LLMs for content moderation, \textit{there is a lack of research in developing approaches to enhance online governance which our work aims to address.}
\section{Preliminaries}

We now detail the communities we examine and the datasets we curate.

\paragraph{Communities:} We categorize communities (subreddits) of interest into two groups for our study:
\textbf{(1) Source subreddits:} These subreddits serve as the foundation for our expert models. We select 7 popular subreddits with a wide spectrum of topics, moderation styles, and community norms\footnote{Although we select these subreddits, our framework can be extended to any other set of subreddits and number of experts at the time of deployment. (See Section \ref{sec:limitations})}: \textit{r/AskHistorians}, \textit{r/AskReddit}, \textit{r/Games}, \textit{r/anime}, \textit{r/changemyview}, \textit{r/politics}, and \textit{r/science}.

\noindent\textbf{(2) Target subreddits:} These subreddits are used for testing the performance and generalization capabilities of \momo{} compared to other baseline approaches. We select 30 diverse subreddits (listed in \autoref{app:subreddits-stats}) chosen for their variety in topics, community sizes, and community norms. 

\paragraph{Datasets:} We curate our data from the publicly available dataset of Reddit comment removals collected between May 10, 2016 and February 4, 2017 by \citet{chandrasekharan_internets_2018}. We create two kinds of datasets for our tasks: 

\noindent\textbf{(1) Community Dataset ($\mathcal{D}_{\text{Community}}$):} This dataset consists of \textit{subreddit}, \textit{comment}, \textit{context}, and \textit{label}, where the \textit{context} is the parent-comment of the original comment, and the \textit{label} is a binary value of `True' or `False' to indicate whether the comment was removed by moderators. $\mathcal{D}_{\text{Community}}$ contains a total of 70,000 entries (7 source subreddits $\times$ 10K comments/subreddit)

\noindent\textbf{(2) Norm Violation Dataset ($\mathcal{D}_{\text{NormVio}}$):} This dataset consists of \textit{norm-violation}, \textit{subreddit}, \textit{comment}, \textit{context}, and \textit{label}, where the \textit{norm-violation} column represents the specific kind of norm the original comment violates or does not violate, as noted by the \textit{label}. We create the labels for this column 
through an LLM-based approach. Specifically, we prompt GPT-4o~\cite{hellogpt4o} with the \textit{context}, \textit{comment}, and the rules of the subreddit and ask it to label each removed \textit{comment} in our source datasets with the \textit{subreddit} rule that it violates. Next, we manually categorize the set of rules from different subreddits into 5 broader norm-violation themes.\footnote{The 5 themes are (1) `Bad Faith or Unsubstantiated Arguments', (2) `Civility and Respect', (3) `Low Effort, Off-Topic, or Non-Substantive Contributions', (4) `Rule Enforcement and Structural Integrity of Discussions', and (5) `Spam, Solicitation, Misinformation, and Machine-Generated Content'.} We then use these mappings as the final label for the \textit{norm-violation} column. Overall, $\mathcal{D}_{\text{NormVio}}$ contains 81,262 entries, balanced across norm-violation categories. The prompt to obtain rule-to-norm violation mappings, and their validity in terms of accuracy ($87\%$ accuracy) and coders' inter-rater reliability (Krippendorff's $\alpha=0.82$) can be found in \autoref{app:dataset-creation}.

\section{MoMoE Framework}

We now explain each component of \momo{} and the rationale behind our choices.

\paragraph{\allocate{}:} Given an incoming context-comment pair, this  operator identifies the appropriate experts within \momo{} and determines their relative importance. We implement two distinct approaches for this allocation: 

\noindent\textbf{(1) Classification-based allocation:} We create an 80:20 train-test split for the source datasets and fine-tune two separate RoBERTa-base model~\cite{liu2019roberta} given the concatenated ``context'' and ``comment'' as input to (i) predict the source \textit{subreddit} label from $\mathcal{D}_{\text{Community}}$ and (ii) predict the \textit{norm-violation} label from $\mathcal{D}_{\text{NormVio}}$. These classifiers for $\mathcal{D}_{\text{Community}}$ and $\mathcal{D}_{\text{NormVio}}$ achieve a test accuracy of 78\% and 62\% respectively.\footnote{See \autoref{app:lora_hyperparams} for fine-tuning hyperparameters.} Next, for allocation, we compute the softmax probabilities from logits of the penultimate layer of the model, and use them directly as expert weights. This approach leverages the model's ability to \textit{identify subreddit-specific linguistic patterns and discussion topics}. 

\noindent\textbf{(2) Similarity-based allocation:} We utilize the SentenceBERT model~\cite{reimers-gurevych-2019-sentence} \texttt{all-mpnet-base-v2} to compute embeddings for all comments in both source and target subreddits. For each comment in the target datasets, we compute two types of averaged cosine similarities: (i) between the embedding of the target comment and the embeddings of all comments in each source subreddit; and (ii) between the embedding of the target comment and the embeddings of all comments in each norm-violation category in the source subreddits. This process yields either: (i) 7 similarity scores (one per source subreddit) or (ii) 5 similarity scores (one per norm-violation category) that each lies in $[-1,1]$. We apply a softmax function ($\tau=0.1$) to convert these scores into probability distributions, which serve as the weights for our experts. This approach captures \textit{semantic similarity between comments across communities}.\footnote{Note that learning allocation weights through backpropagation is another alternative which we do not explore in this work as one of our key goals is transparent allocations.}

\paragraph{\predict{}:} This operator is the core component of \momo{} that uses the mixture-of-experts inspired framework to determine moderation outcomes. This component takes the context-comment pair as input and produces binary moderation decisions from multiple specialized experts. For our expert models, following~\citet{zhan-etal-2025-slm}, we leverage two state-of-the-art open-source small language models (SLMs): \textit{Llama-3.1-8B}~\cite{dubey2024llama}, and \textit{Mistral-Nemo-Instruct}~\cite{mistralMistralNeMo}. Each model is fine-tuned using Low-Rank Adaptation (LoRA)~\cite{hu2021lora} to create specialized experts for content moderation. We fine-tune these models using rule-based prompting. The detailed prompts used for LoRA fine-tuning as well as hyperparameter details can be found in \autoref{app:lora_hyperparams}. We explore two distinct approaches to expert specialization: 

\noindent\textbf{(1) Community-based Experts}: This approach (\momoco{}) creates subreddit-specific experts by fine-tuning SLMs on data from each source subreddit. Using an 80:20 train-test split of $\mathcal{D}_{\text{Community}}$ stratified by \textit{subreddit}, we fine-tune separate expert models for each source subreddit. Each expert specializes in the specialized rules and moderation patterns of its respective community.

\begin{table*}[t]
\small
\sffamily
\centering
\resizebox{0.9\linewidth}{!}{
\begin{tabular}{lcccccccccc}
\hline
& \multicolumn{2}{c}{\textbf{LLM}} 
& \multicolumn{2}{c}{\textbf{Base SLM}} 
& \multicolumn{2}{c}{\textbf{Fine-tuned SLM}} 
& \multicolumn{2}{c}{\textbf{\momoco{}}} 
& \multicolumn{2}{c}{\textbf{\momonv{}}} \\
\cmidrule(lr){2-3}\cmidrule(lr){4-5}\cmidrule(lr){6-7}\cmidrule(lr){8-9}\cmidrule(lr){10-11}
\rowcolor{blue!10}\textbf{Subreddit} 
& \textbf{GPT-4o-mini} & \textbf{GPT-4o} 
& \textbf{Llama} & \textbf{Mistral} 
& \textbf{Llama} & \textbf{Mistral} 
& \textbf{Llama} & \textbf{Mistral}
& \textbf{Llama} & \textbf{Mistral} \\
\hline
r/anime          & \cellcolor{red!40}41.8 & \cellcolor{red!40}33.0 & \cellcolor{red!40}56.6 & \cellcolor{red!40}12.6 & 63.1 & 75.5 & \cellcolor{pink!25}61.4 & \cellcolor{pink!25}72.9 & \cellcolor{red!25}58.7 & \cellcolor{red!25}70.6 \\
r/AskHistorians  & \cellcolor{red!40}26.3 & \cellcolor{red!40}38.2 & \cellcolor{red!40}54.9 & \cellcolor{red!40}8.3  & 67.4 & 76.9 & \cellcolor{pink!25}66.9 & \cellcolor{pink!25}74.5 & \cellcolor{red!25}63.6 & \cellcolor{red!25}72.6 \\
r/AskReddit      & \cellcolor{red!40}51.7 & \cellcolor{red!40}51.5 & \cellcolor{green!25}56.3$^\ast$ & \cellcolor{red!40}40.6 & 55.3 & 62.5 & \cellcolor{green!25}55.7$^\ast$ & \cellcolor{pink!25}60.8 & \cellcolor{red!25}49.7 & \cellcolor{pink!25}60.1 \\
r/changemyview   & \cellcolor{red!40}79.4 & \cellcolor{red!40}74.9 & \cellcolor{red!40}57.7 & \cellcolor{red!40}55.4 & 90.3 & 91.8 & \cellcolor{red!25}84.8 & \cellcolor{red!25}86.7 & \cellcolor{red!25}83.5 & \cellcolor{red!25}85.4 \\
r/Games          & \cellcolor{red!40}45.7 & \cellcolor{red!40}44.2 & \cellcolor{red!40}55.6 & \cellcolor{red!40}22.9 & 69.9 & 74.3 & \cellcolor{green!25}70.3$^\ast$ & \cellcolor{pink!25}72.4 & \cellcolor{pink!25}68.4 & \cellcolor{red!25}71.5 \\
r/politics       & \cellcolor{red!25}71.8 & \cellcolor{red!25}72.2 & \cellcolor{red!40}54.3 & \cellcolor{red!40}53.8 & 72.5 & 78.1 & \cellcolor{green!25}72.8 & \cellcolor{green!25}78.3 & \cellcolor{pink!25}70.2 & \cellcolor{red!25}73.6 \\
r/science        & \cellcolor{red!40}42.9 & \cellcolor{red!25}66.6 & \cellcolor{red!40}52.2 & \cellcolor{red!40}30.2 & 63.0 & 72.7 & \cellcolor{pink!25}62.2 & \cellcolor{red!25}69.4 & \cellcolor{pink!25}61.3 & \cellcolor{red!25}67.4 \\
\hline
\end{tabular}}
\caption{\textbf{MoMoE Performance on Source Subreddits.}  
Micro-F1 scores (higher is better) are colored by relative drop vs.\ the corresponding fine-tuned SLM (for \momoco{}, \momonv{}, and Base SLM) or vs.\ the best fine-tuned SLM (for LLMs):  
\colorbox{pink!25}{\(\le\) 2.5 \% drop}, \colorbox{red!25}{2.5–10 \% drop}, \colorbox{red!40}{\(>\) 10 \% drop}, and \colorbox{green!25}{improvement} \small(${}^\ast$$p<0.05$). The \momo{} models use the classification-based strategy and dot-product based aggregation.}
\label{tab:MoMoE-results}
\end{table*}

\noindent\textbf{(2) Norm-violation Experts:} This approach (\momonv{}) creates subreddit-specific experts by fine-tuning SLMs on data from each norm-violation category. Using an 80:20 train-test split of $\mathcal{D}_{\text{NormVio}}$ stratified by \textit{norm-violation}, we fine-tune separate expert models for each of the $5$ categories, where each expert is specialized in detecting particular kinds of norm violations. 

These complementary approaches offer distinct advantages for content moderation. We hypothesize that \textit{community-based experts would excel at capturing the nuanced, community-specific norms} that may vary significantly across different subreddits. In contrast, \textit{norm-violation experts should generalize better across communities} by focusing on fundamental categories of problematic content that tend to be universally unacceptable across most online spaces, albeit to possibly varying extents.

\paragraph{\aggregate{}:} This operator is responsible for combining the decisions of multiple experts using their allocated weights to produce a final outcome. We implement this component with a ``Top-K'' approach that selects the K experts with the highest allocation weights. Within this framework, we explore two distinct aggregation strategies: 

\noindent\textbf{(1) Dot Product:} We compute a weighted composition score by taking the dot product between two vectors of dimension K: (i) a binary decision vector from the experts; and (ii) the allocation weight vector determined by the \allocate{} operator.  We apply a threshold at $0.5$---if the composition score exceeds this threshold, it returns `True' (comment should be removed); otherwise, it returns `False'.

\noindent\textbf{(2) Majority Vote:} We determine the final outcome by taking a simple majority vote among the Top-$K$ chosen experts. The decision supported by more than half of the experts is final. 

These aggregation strategies allow us to evaluate the trade-offs between \textit{relying on a few highly relevant experts versus incorporating a broader consensus}, and different aggregation methods.

\paragraph{\explain{}:} This operator is the final component of MoMoE, aimed at using a strong LLM like GPT-4o for providing transparent justifications for moderation decisions to human moderators who would use such a framework in practice (See \autoref{app:explanation} for prompt design). We generate explanations that detail the reasoning behind MoMoE's decisions.

These explanation strategies have many benefits. First, it enables moderators to easily understand which experts were most relevant to a particular moderation decision and why. Second, it helps moderators identify the specific types of norm violations or community standards that were considered while making these decisions, which helps in facilitating more consistent and fair moderation decisions across similar cases that may have otherwise been treated differently or overlooked. Most importantly, the transparent nature of these explanations allows moderators to identify potential biases or miscalibrations within the MoMoE framework. 

\paragraph{Evaluation Baselines:} \textbf{(A) Performance:} We evaluate MoMoE against the following baselines: \textbf{(1) Detoxify}~\cite{Detoxify}: a model that computes toxicity and is thresholded at $0.5$ to determine the moderation outcome. \textbf{(2) Global SLMs:} Small language models fine-tuned separately on entire train-split of $\mathcal{D}_{\text{Community}}$ (G-Llama$_\text{Community}$ and G-Mistral$_\text{Community}$) and $\mathcal{D}_\text{NormVio}$ (G-Llama$_\text{NormVio}$ and G-Mistral$_\text{NormVio}$). \textbf{(3) Global LLMs:} Zero-shot prompted LLMs (GPT-4o and GPT-4o-mini) to predict whether a comment will be removed. Note that we do not perform few-shot ICL as prior work~\cite{zhan-etal-2025-slm} has shown that it does not reliably improve performance due to lack of ways to incorporate examples relevant to specific communities. \textbf{(B) Explainability:} We perform manual validation to check that the generated explanations reliably reflect the course of \momo{}'s decision-making trace.
\section{Results}

We now evaluate \predict{} in terms of Micro-F1 (F1 hereafter), highlighting the key tradeoffs compared to community-specific SLMs and LLMs, and demonstrate the benefits of \momo{} operators.

\begin{figure*}[t]
    \centering
    \includegraphics[width=\linewidth]{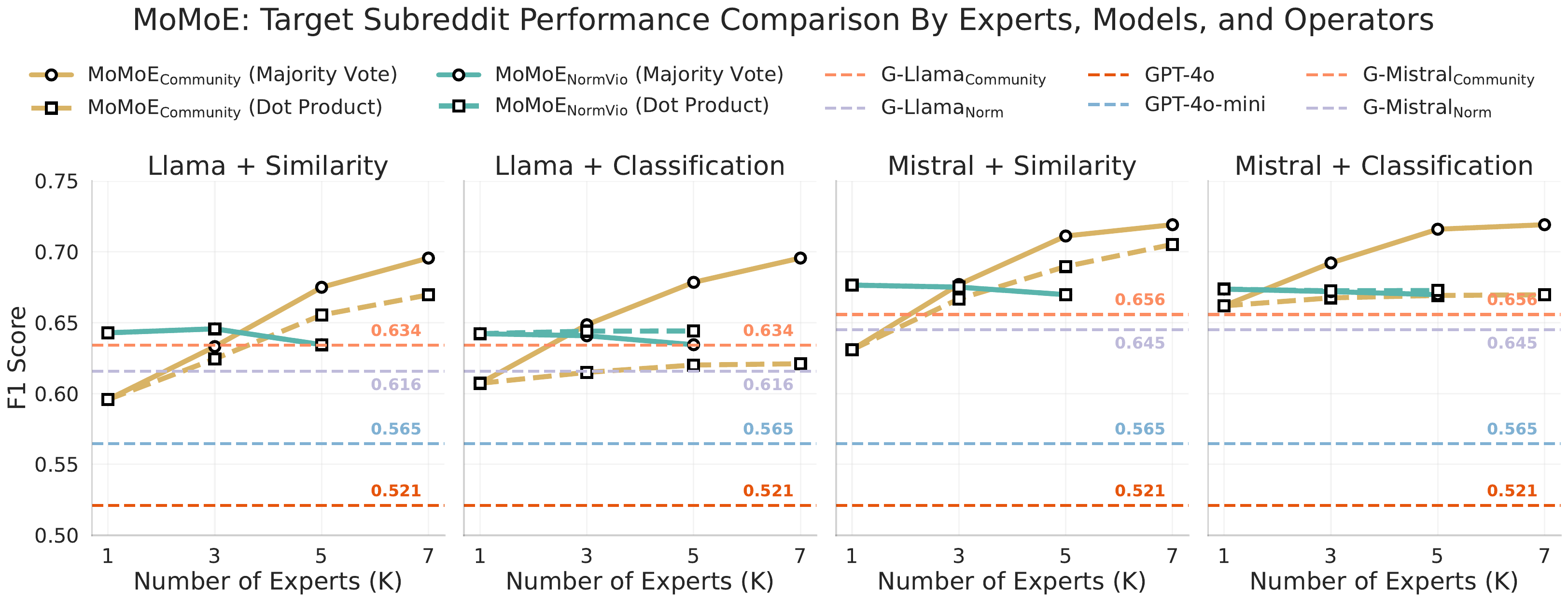}
    \caption{Performance of MoMoE on Target Subreddits reveals that both \momoco{} and \momonv{} perform competitively against baselines either matching or outperforming them in terms of Micro-F1 score.}
    \label{fig:momoe_all_facets}
\end{figure*}

\paragraph{\momo{} Performance on Source Subreddits:} We first evaluate how \momoco{} and \momonv{} perform on the test splits of the source subreddits, the train-data of which their experts were fine-tuned on. This evaluation acts as a sanity check to ensure that shifting from community-specific SLMs to a mixture-of-moderation-experts does not lead to a large drop in performance and at the same time can still outperform off-the-shelf SLMs and LLMs. From \autoref{tab:MoMoE-results}, we see that both \momo{} approaches show only moderate performance drops compared to community-specific fine-tuned SLMs, with \momoco{} experiencing smaller drops (typically $\le$2.5\%) than \momonv{} across most subreddits. All drops are significant by Welch's t-test~\cite{welch1947generalization} for \momo{} ($p$<0.05) and LLMs ($p$<0.001). Notably, \momoco{} even outperforms the source SLMs in two cases ($p$<0.05), highlighting that the ensemble approach can sometimes benefit from shared moderation patterns across communities. Furthermore, both \momo{} variants substantially outperform off-the-shelf LLMs GPT-4o-mini and GPT-4o, which show dramatic performance drops (often >10\%) compared to specialized SLMs. This shows that \momo{} performs well on source data, encouraging us to apply \momo{} to a more challenging setting where we adopt a community-agnostic approach without prior knowledge of comment origins.

\paragraph{MoMoE Performance Against Baselines:} 

In our subreddit-agnostic setting where we use comments from 30 target subreddits, \autoref{fig:momoe_all_facets} demonstrates that \momo{} consistently matches or outperform baselines. The strongest baselines are the Global SLMs G-Mistral$_\text{Community}$ and G-Llama$_\text{Community}$ with F1 scores of 0.656 and 0.634 respectively, while the LLMs GPT-4o and GPT-4o-mini show much weaker performance with F1 scores of 0.521 and 0.565 respectively. The toxicity classifier based on Detoxify (not plotted) showed the worst performance with an F1 score of 0.38. The Mistral-based \momoco{} with both classification and similarity allocation strategies with majority-voting based aggregation is the best performing configuration with F1 scores of 0.72. All improvements for \momoco{} against the strongest baseline---community-based Global SLMs---are significant at K=5 and 7 ($p<0.001$), except dot-product aggregation for Llama with classification allocation. For \momonv{}, all improvements at K=1 and K=3, and at K=5 only improvements by Mistral, are significant ($p<0.05$).

We find that for \momoco{}, increase in the number of experts $K$ generally leads to a notable increase in performance, while for \momonv{}, increasing the number of experts maintains or slightly drops performance. Based on our rule-to-norm mappings (\autoref{app:dataset-creation}), we hypothesize that in the case of norm-violation experts even if one of the experts deems the comment as violating, it should be removed as the violation of any of these norms is harmful across most communities. Incorporating more experts could therefore lead to misclassification in some cases. We expand on these findings with a precision-recall trade-off analysis comparing \momoco{} and \momonv{} in \autoref{app:pr-tradeoff}. Also see \autoref{app:imbalanced-results} for performance of \predict{} on imbalanced test-split in terms of AUC and Macro-F1.

\paragraph{Impact of Operator Choice:} In terms of the \allocate{} operators, for low number of experts (K = 1 or 3) classification-based allocation slightly outperforms similarity-based allocation ($\approx0.04$ F1 on average, $p<0.05$), while for higher number of experts (K = 5 or 7), both allocation strategies are equivalent with negligible differences in F1 scores. In terms of \aggregate{} operators, we find that for \momoco{}, majority voting consistently outperforms dot-product based aggregation. On the other hand, for \momonv{} the two aggregation strategies are roughly equivalent with dot product slightly outperforming majority vote in the case of classification-based allocation ($\approx0.01$ F1 on average, $n.s.$). We discuss potential reasons for some of these observations in Section \ref{sec:further-analysis}. 

\begin{figure}[t]
    \centering
    \includegraphics[width=\linewidth]{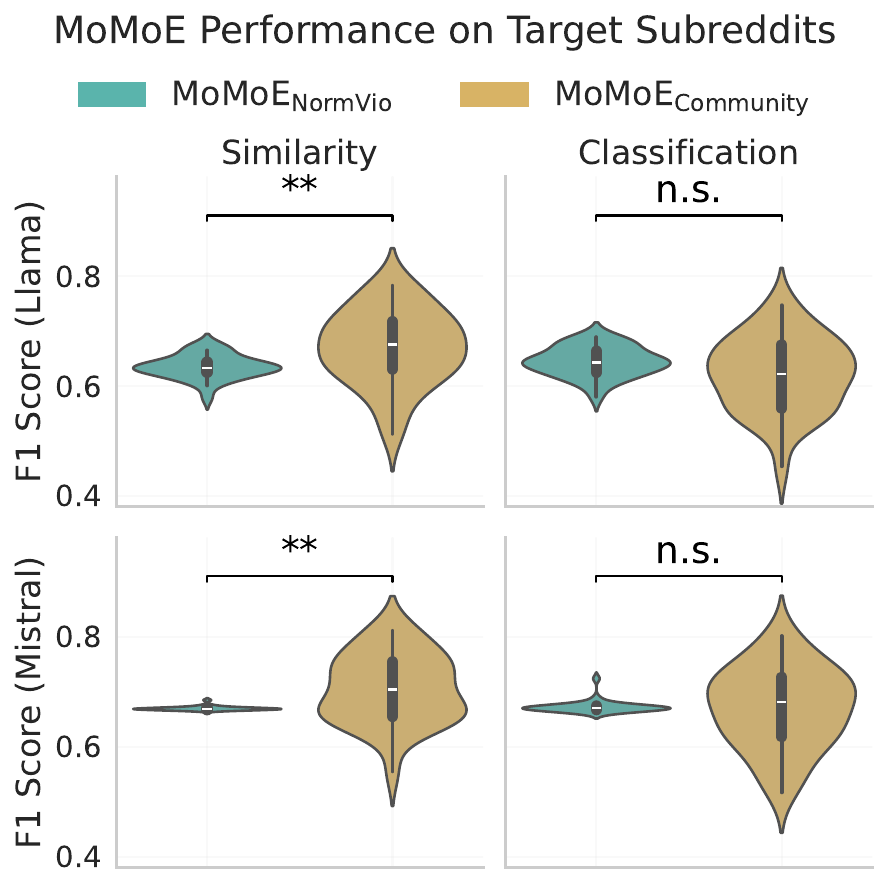}
    \caption{Comparing F1 score performance with dot-product based aggregation we observe that while \momoco{} ~provides a wider range of performance across subreddits ($\approx0.45-0.8$), \momonv{} ~gives consistent moderate performance across subreddits ($\approx0.65$). \small(${}^\ast$$p<0.05$, ${}^{\ast\ast}$$p<0.01$, ${}^{\ast\ast\ast}$$p<0.001$)}
    \label{fig:momoe_f1_scores}
\end{figure}

\paragraph{Dissecting Performance of \predict{} on Target Subreddits:}

We now dive deeper into the performance of Llama-based \momo{} on the target subreddits, providing insights into the key differences between \momoco{} versus \momonv{} with the dot-product based aggregation strategy for this case study, although these trends are consistent with majority-voting as well. All significance levels are from Welch's $t-$test~\cite{welch1947generalization}.

\autoref{fig:momoe_f1_scores} shows key differences in performance variability between our two \momo{} approaches. The Llama-based \momoco{} demonstrates a much wider performance range across target subreddits, with F1 scores spanning from as low as $0.45$ for \textit{r/hillaryclinton} with the classification allocation strategy to as high as $0.78$ for \textit{r/Overwatch} with the similarity allocation strategy. Overall, \momoco{} with similarity-based allocation achieves a mean F1 score of $0.67~(\pm0.07)$, while with classification allocation it achieves $0.62~(\pm0.07)$ ($p<0.01$). In contrast, \momonv{} shows consistent performance across all subreddits, with a much narrower range of F1 scores from $0.58$ for \textit{r/pokemontrades} with similarity-based allocation to $0.69$ for \textit{r/DestinyTheGame} with classification-based allocation. For \momonv{}, both allocation strategies yield similar overall performance with mean F1 scores of $0.64~(\pm0.02)$ for similarity- and $0.64~(\pm0.03)$ for classification-based allocation ($n.s.$). Mistral \momo{} shows very similar trends (See \autoref{app:mistral-spread}).

\paragraph{How do \allocate{} and \aggregate{} operators affect outcomes of \predict{}?}\label{sec:further-analysis}

In principle, the \allocate{} operator is similar in functionality to a \textit{jury allocator} in the jury learning setting~\cite{gordon_disagreement_2021} by helping identify which experts and in what proportion should determine \momo{}'s prediction. As a result, a natural followup question after observing the strong performance of \momo{} is analyzing what kind of expert compositions are facilitated by the two strategies of classification- and similarity-based allocation. One notable thing about our framework is that if any single expert is allocated a weight of more than $0.5$, the decision taken by that expert would be the final one, and therefore in such cases only that expert is required to arrive at an outcome.

We find that since classification-based allocation is essentially a prediction task, on average for \momoco{} and \momonv{}, $72.7\%$ and $69.3\%$ of all predictions on the target subreddits are guided by just one expert respectively. For \momoco{} the most and least solely-utilized experts were \textit{r/AskReddit} ($23.1\%$ of cases) and \textit{r/AskHistorians} ($3.2\%$ of cases), while for \momonv{} they were the \textit{`Civility and Respect'} expert ($38.5\%$ of cases) and \textit{`Bad Faith or Unsubstantiated Arguments'} expert ($2.8\%$ of cases). Similarity-based allocation on the other hand shows the opposite pattern, where at least 3 experts are needed in $87\%$ of all predictions, and the allocation weights are much more uniform in comparison to classification-based allocation.

These insights also provide an explanation for the performance differences between majority voting and dot product based \aggregate{} operators. Specifically for \momoco{}, we see in \autoref{fig:momoe_all_facets} that majority vote aggregation performs very similarly across both similarity- and classification-based allocation, as despite the difference in allocation distributions, the top experts remain similar across strategies. On the other hand, dot product aggregation is more sensitive to allocation distributions and as a result we see a clear drop in F1 performance from similarity-based to classification-based allocation by $\approx0.05$ and $\approx0.04$ for all experts with Llama and Mistral respectively. This further indicates that with dot-product based aggregation, our ensemble framework can leverage knowledge from multiple experts to get large benefits over a single, dominant expert.

\section{Error Analysis of \momo{} Decisions} 

In order to identify and analyze potential failure cases of the moderation decisions made by \momo{}, we perform an error analysis. Understanding of failure modes is crucial to improve the system and deploy it safely in real-world settings.

Specifically, we analyzed results on the target community \textit{r/movies} for the best-performing setting: Mistral-based \momoco{} with classification-based allocation and majority vote based aggregation. We sample 100 instances where \momo{} got the moderation outcome incorrect (50 False Positives and 50 False Negatives) to perform a qualitative error analysis, open-coding each error instance into fine-grained categories.

\begin{table}[t]
\small
\sffamily
\centering
\resizebox{\linewidth}{!}{
\begin{tabular}{l r}
\rowcolor{blue!10}\multicolumn{2}{c}{\textbf{False Positives}}\\
\rowcolor{gray!10}\textbf{Error Category} & \textbf{\# Instances} \\
Sarcasm and Jokes & 16 \\
Civil Disagreement / Critique & 7 \\
Pop Culture / Meme Reference & 9 \\
Innocuous Factual Statement & 7 \\
Self-Deprecation / Light-weight Banter & 3 \\
Ambiguous / Need More Context & 8 \\
\addlinespace
\rowcolor{blue!10}\multicolumn{2}{c}{\textbf{False Negatives}}\\
\rowcolor{gray!10}\textbf{Error Category} & \textbf{\# Instances} \\
Not-so-civil Disagreement & 10 \\
Missed Hate Speech and Slurs & 8 \\
Social Stereotypes including Sexism & 9 \\
Endorsement of Illegal Piracy & 11 \\
Off-topic Comments & 12 \\
\hline
\end{tabular}}
\caption{\textbf{Error Taxonomy Summary.} Prevalence of representative patterns in false positives (FP) and false negatives (FN) from the manual audit of \textit{r/movies}.}
\label{tab:error-taxonomy}
\end{table}

From \autoref{tab:error-taxonomy}, we find that most false positives arise from sarcasm, jokes, and innocuous or civil disagreement-cases where tone and intent are playful or critical but non-toxic. False negatives on the other hand cluster around not-so-civil disagreement, nuanced hate speech/stereotypes, and policy-violating yet context-light content (e.g., piracy or off-topic posts). This pattern indicates that the model often over-penalizes borderline or humorous language without sufficient pragmatic/contextual cues, yet still under-detects subtle toxicity and norm violations that require fine-grained semantic and socio-linguistic reasoning.

These errors suggest a need for stronger contextualization in the form of conversation/thread history, community context, and better modeling of pragmatic phenomena such as sarcasm, stance, politeness. While the NLP community has long targeted these capabilities, our results underscore that distinguishing civil versus toxic disagreement and detecting nuanced hate speech remain open challenges, warranting future work in this area.

\section{\momo{} Explanations}

The final component of \momo{} is the \explain{} operator that turns raw model signals into concise, moderator‑facing rationales. The operator is designed around three key principles inspired by HCI research in human-AI collaborative systems: \textbf{(i) Progressive Disclosure:} provide a one‑line verdict first and allow cascading when needed~\cite{choi_convex_2023}; \textbf{(ii) Reliability:} the explanation is based on the same evidence driving the decision~\cite{selbst_fairness_2019}; and
\textbf{(iii) Actionability:} by surfacing disagreements or low confidence so that
moderators know when to intervene~\cite{koshy2024venire}. 

During inference, we log a trace of every expert containing its vote and confidence. Next, we prompt GPT‑4o to convert the trace into a three‑level JSON explanation with keys \textbf{Summary}, \textbf{Key Points}, and \textbf{Trace}. The \textit{Summary} provides a simple actionable insight with a `Remove', 'Keep', 'Review' decision, a brief reason inferred from the top expert, and level of 'High' or 'Low' consensus among experts. The \textit{Key Points} provides information about the top expert along with its allocation, and details about consensus on the decision. The \textit{Trace} represents the original trace for audit including the decision and \momo{} confidence-level, along with LLM-inferred salient spans that could possibly indicate problematic areas in the comment. See \autoref{app:explanation} for the prompts, in which we include three‑shot exemplars and query GPT‑4o with \textit{temperature}=0 to generate explanations.

\begin{tcolorbox}[width=.48\textwidth, colframe=darkblue]
\small \textbf{Summary:} Review: Hate Speech; Low Consensus\\
\textbf{Key Points:}\\
1. Top expert: `Civility and Respect' (0.35)\\
2. Low consensus: 2/5 experts – Review\\
\textbf{Trace:}\\
1. Decision: ``REMOVE''\\
2. Confidence: 0.58\\
3. Salient Spans: [``go back to your country'', ``lmao'']
\end{tcolorbox}

We manually sample four explanation samples from each target subreddit totaling 120 examples, and we obtain a 100\% reliability, which indicates that given \momo{}'s trace, GPT-4o can generate nearly perfect explanations in terms of reliably reflecting the decision processes. The \explain{} operator is therefore model-agnostic, and enhances transparency without overwhelming moderators.

\section{Discussion and Implications}

\paragraph{Cross-community Content Moderation:} \momo{} improves cross‑community moderation by pooling knowledge from a small set of specialized experts instead of fine-tuning one model per subreddit. Its architecture leverages the \allocate{} and \aggregate{} operators adapt to unfamiliar communities while retaining high F1 on known ones. This design lowers the data barrier for new or low‑resource communities and shows that performance need not be sacrificed for good generalization capabilities. For NLP researchers, our results show the power of lightweight expert ensembles without resorting to generalist models, highlighting the need for research on efficient transfer and dynamic selection of experts.

\paragraph{Complementary Strengths of `Community' and `Norm-violation' Experts:} With the \predict{} operator, we show that community experts can provide a wider range of performance across unseen communities while norm-violation experts provide consistently strong performance. This shows that while a community ensemble can prove beneficial when the target community has similar content or norms to the source communities, they may struggle to adapt to completely different kinds of communities. An ensemble based on broader norm-violations, on the other hand, may offer consistent performance, as research has shown that such violations are often shared across communities~\cite{chandrasekharan_internets_2018}. This suggests that content deemed undesirable in one community is also likely to be considered norm-violating in others.

\paragraph{AI-assisted Content Moderation:} Beyond raw accuracy, \momo{} illustrates a practical blueprint for AI‑assisted moderation. The \explain{} operator converts model traces into layered, human‑readable rationales that surface confidence and expert disagreement, giving moderators clear cues about when to intervene. This workflow shifts the narrative from \textit{automation} to \textit{augmentation}, letting moderators handle edge cases while the system filters routine decisions. For HCI researchers, the framework highlights the value of progressive disclosure and actionable transparency in sociotechnical tools, opening avenues for studying trust calibration and interface design in mixed‑initiative governance systems, and also real-time deployment and user-studies using \momo{}.

\paragraph{Directions for future work:} Although our offline metrics are promising, they provide an incomplete view of moderator support. Systems such as \momo{} should be assessed with user-centric evaluations that foreground human outcomes~\cite{selbst_fairness_2019}. Useful study designs could include within-subjects experiments to assess outcomes such as decision quality, time-to-decision, workload, reliance, and perceived transparency and trust. 

Further, moderation systems should not focus solely on removals, but also on supporting \emph{proactive} approaches that elevate identify and reinforce desirable content~\cite{grimmelmann_virtues_2015, lambert_positive_2024}. One direction is to add ``prosocial experts'' trained to detect prosocial~\cite{bao_conversations_2021} and high-quality discourse~\cite{goyal2025language,lambert2025mind}, enabling ranking/boosting policies alongside removal recommendations.
\section{Conclusion}

We introduce \momo{}, a modular ensemble framework that scales content moderation of online communities beyond resource-intensive community-specific approaches in a transparent manner. \momo{} attains strong F1 scores on 30 unseen communities, matching or surpassing fine-tuned SLM baselines and comprehensively outperforming zero-shot LLMs. \momo{} explanations turn raw decision traces into concise, moderator-ready rationales that were judged reliable in manual inspection. These findings highlight expert ensembles as a viable path toward data-efficient, human-AI collaborative governance. We outline directions for future work on adaptive expert selection, real-time user studies, and deployment of moderation systems at scale.

\section{Limitations}\label{sec:limitations}

Our work has limitations, which also suggest interesting future directions. 

\paragraph{(1) Specific choice of in-domain subreddits:} We evaluate \momo{} on seven subreddits that were deliberately varied in size, topic, and rule complexity, but we do not test how alternative in-domain sets influence performance. Other community selections or a different mix or count of experts might improve or lower the performance we observe. Because our primary goal was to establish \momo{} as a viable framework, we chose breadth over exhaustive tuning; the reported results therefore serve as a proof-of-concept. Future work should replicate our study with additional subreddit collections and systematically vary the number and granularity of both community and norm-violation experts.

\paragraph{(2) No exploration of multi-agent LLM frameworks:} We deliberately restrict \momo{} to lightweight, single-pass SLM experts rather than a multi-agent setup in which several large LLMs interact or debate. This choice was informed by recent work that found that fine-tuned SLMs surpass zero- and few-shot LLMs in moderation while remaining far cheaper to deploy~\cite{zhan-etal-2025-slm}, and it keeps our design focused on data-efficient generalization to unseen communities rather than a multi-agent orchestration. Future work could build human-AI collaborative, multi-agent systems where each agent embodies a community, a moderator persona, or a norm-violation category.

\paragraph{(3) Label noise and annotation bias:}
Our training and test labels are derived from ground truth moderator actions collected by prior work, which can be inconsistent and influenced by local norms, human biases, or fatigue. This noise may both inflate and depress measured performance. Similarly, while our LLM-based approach for constructing $\mathcal{D}_\text{NormVio}$ shows high accuracy and inter-annotator agreement, it is imperfect which could lead to some amount of performance drop.

\paragraph{(4) English-only evaluation:}
All subreddits in our study are English-speaking. \momo{}’s experts and especially the \explain{} operator therefore rely on English language cues. Generalizing to multilingual or code-switched communities will likely demand new experts and prompt adaptation, and is something that future work could explore.

\paragraph{(5) Latency and resource overhead:}
Although each of our experts is a lightweight $4-$bit quantized SLM, invoking multiple experts plus GPT-4o for explanations adds latency and compute relative to a single classifier. While this is not an issue for deployment as such systems would likely be hosted on a Reddit backend, in high-traffic settings this could raise deployment costs.

\paragraph{(6) Lack of user-centric evaluation:}
We measure explanation quality with manual validation but do not study how \momo{} affects actual moderation workflows on Reddit, and the trust or decision time of Reddit moderators. Controlled user studies and longitudinal field deployments are needed to validate practical utility and uncover such findings.
\section*{Ethical Considerations}

\momo{} is targeted at the reduction of harmful content, yet its deployment could raise several ethical questions.  First, moderation labels inherited from Reddit may encode community-specific biases. We mitigate this by releasing our code and allowing researchers to audit or retrain experts on alternative annotations. Second, false positives in moderation can censor legitimate speech while false negatives can expose community users to harm, and therefore we design the Explain operator to surface confidence and disagreement so that humans remain in the loop for contentious cases. Finally, since we use GPT-4o-mini and GPT-4o we ensure to comply with the OpenAI API's terms of use policies.\footnote{\href{https://openai.com/policies/terms-of-use/}{https://openai.com/policies/terms-of-use/}} We believe that our transparent reporting of limitations, along with the open release of artifacts upon publication will ensure that we minimize introducing any new harms.

\section*{Acknowledgments}

A.G. was supported by compute credits from the OpenAI Researcher Access Program. This work used the Delta system at the National Center for Supercomputing Applications through allocation \#240481 from the Advanced Cyberinfrastructure Coordination Ecosystem: Services \& Support (ACCESS) program, which is supported by National Science Foundation grants \#2138259, \#2138286, \#2138307, \#2137603, and \#2138296.

The authors thank the members of the Social Computing Laboratory (SCUBA) at the University of Illinois and anonymous reviewers of the ACL Rolling Review for their insightful suggestions that helped improve the work.

\bibliography{references}

\appendix

\section{Target Subreddits}\label{app:subreddits-stats}

In this section, we list the 30 subreddits we used as our target communities for downstream evaluation of \momo{}. These communities were randomly sampled from the original comment removal dataset released by \citet{chandrasekharan_crossmod_2019} after removing the source communities our in-domain experts were fine-tuned on. Sampling these subreddits from the original dataset was crucial as we needed ground truth removal labels for evaluation.

r/food, r/PoliticalDiscussion, r/hearthstone, r/OldSchoolCool, r/gonewild, r/spacex, r/WTF, r/pokemongo, r/DestinyTheGame, r/BlackPeopleTwitter, r/nottheonion, r/Overwatch, r/pokemontrades, r/explainlikeimfive, r/IAmA, r/personalfinance, r/hillaryclinton, r/news, r/leagueoflegends, r/funny, r/toronto, r/depression, r/pcmasterrace, r/OutOfTheLoop, r/HistoryPorn, r/ShitRedditSays, r/asoiaf, r/relationships, r/nba, r/movies.

\section{Norm Violation Dataset Creation}\label{app:dataset-creation}

As outlined in the main text, we use an LLM-based approach to create our norm-violations dataset $\mathcal{D}_\text{NormVio}$ using GPT-4o. We use the prompt below asking the LLM to classify each instance of rule-violating comment into a specific set of rules it violated. We then augment this ``positive'' class with a sample of non-violating comments to form a balanced dataset for fine-tuning experts with an 80\% split, while the rest is used for testing.

\begin{tcolorbox}[
  title=\textbf{PROMPT},
  colback=gray!10,
  colframe=gray!60!black,
  fonttitle=\bfseries,
  fontupper=\ttfamily\small,
  rounded corners,
  breakable 
]
You are an expert moderator for the r/\{SUBREDDIT\} subreddit on Reddit. \\
Here is a description of the subreddit: \{SUBREDDIT\_DESCRIPTION\}

\noindent You are given a comment, the preceding context which it is replying to, and a list of rules for the subreddit. \\
This comment was removed by the moderators of the subreddit, and your task is to determine which rule(s) the comment violates.

\medskip
\textbf{Context:} \\
\{CONTEXT\}

\medskip
\textbf{Comment:} \\
\{COMMENT\}

\medskip
\textbf{Rules:} \\
\{RULES\}

\medskip
Please return the list of rule number(s) that the comment violates in a list format: e.g., 5, 7, 9. \\
If the comment violates only one rule, return a list with one element: e.g., 9. \\
Even if you think the comment violates no rules or you are not sure, return the rule it is most likely to violate and nothing else.
\end{tcolorbox}

Once the LLM classified comments into specific rules, we then manually grouped the rules from all in-domain subreddits in broader norm-violation categories, ending up with $5$ themes listed below. To ensure annotation quality, we have the first and second authors manually validate a sample of $140$ comments ($20$ from each source subreddit) obtaining an accuracy of $87\%$, and an inter-rater reliability (IRR) Krippendorff’s $\alpha=0.82$ which denotes high agreement~\cite{hayes2007answering}.

\begin{tcolorbox}[
  title=\textbf{Norm-Violation Mapping},
  colback=blue!8,
  colframe=blue!60!black,
  fonttitle=\bfseries,
  rounded corners,
  breakable,
  fontupper=\sffamily\small,
]

\setlist[itemize]{noitemsep, topsep=1pt, leftmargin=0.5cm}
\setlist[itemize,2]{label=--, leftmargin=0.5cm} 

\textbf{Civility and Respect}
  \begin{itemize}
    \item \textit{r/science}: 2.\ No abusive or offensive comments
    \item \textit{r/politics}: 3.\ No incivility or personal attacks towards users
    \item \textit{r/politics}: 4.\ No flaming, baiting, or trolling
    \item \textit{r/AskReddit}: 2.\ Be respectful to other users at all times
    \item \textit{r/AskHistorians}: 1.\ Users shall behave with courtesy
    \item \textit{r/changemyview}: 2.\ No rude/hostile comment
    \item \textit{r/AskHistorians}: 9.\ No racist or bigoted comments
    \item \textit{r/politics}: 1.\ No hateful speech
    \item \textit{r/Games}: 2.\ No attacks / witch-hunts / bigotry / inflammatory language
  \end{itemize}

\textbf{Low-Effort, Off-Topic, or Non-Substantive Contributions}
  \begin{itemize}
    \item \textit{r/science}: 1.\ No off-topic comments, memes, low-effort jokes
    \item \textit{r/Games}: 3.\ No off-topic or low-effort content
    \item \textit{r/anime}: 2.\ No memes, reaction images, shitposts
    \item \textit{r/anime}: 1.\ Everything posted must be anime-specific
    \item \textit{r/Games}: 1.\ No content primarily for humor or entertainment
    \item \textit{r/changemyview}: 5.\ No comment that doesn’t contribute meaningfully
    \item \textit{r/AskHistorians}: 2.\ Comments must be in-depth and comprehensive
  \end{itemize}

\textbf{Bad-Faith or Unsubstantiated Arguments}
  \begin{itemize}
    \item \textit{r/science}: 3.\ Non-professional personal anecdotes will be removed
    \item \textit{r/science}: 4.\ Criticism must assume basic competence of researchers
    \item \textit{r/science}: 5.\ Dismissing established findings must provide evidence
    \item \textit{r/AskHistorians}: 3.\ Comments should reflect topic familiarity
    \item \textit{r/AskHistorians}: 4.\ No speculative or anecdotal comments
    \item \textit{r/changemyview}: 3.\ No bad-faith accusation
    \item \textit{r/anime}: 4.\ Do not post heavily NSFW content
  \end{itemize}

\textbf{Spam, Solicitation, Misinformation, Machine-Generated Content}
  \begin{itemize}
    \item \textit{r/politics}: 2.\ No spam or solicitation
    \item \textit{r/AskReddit}: 4.\ No spam, machine-generated content, or karma farming
    \item \textit{r/anime}: 5.\ No spam, low-effort comments, or karma farming
    \item \textit{r/AskHistorians}: 6.\ Comments should not be only links or quotations
    \item \textit{r/AskHistorians}: 7.\ Must cite all quotes; no plagiarism
    \item \textit{r/AskHistorians}: 8.\ Comments should not consist solely of jokes
    \item \textit{r/AskReddit}: 3.\ No harmful misinformation
  \end{itemize}

\textbf{Rule Enforcement and Structural Integrity of Discussions}
  \begin{itemize}
    \item \textit{r/AskReddit}: 1.\ No begging for goods, services, or awards
    \item \textit{r/anime}: 3.\ Do not link to illegal content
    \item \textit{r/AskHistorians}: 5.\ No political agendas or moralising
    \item \textit{r/changemyview}: 4.\ No delta abuse or misuse
    \item \textit{r/changemyview}: 1.\ Top-level comments must challenge OP
    \item \textit{r/science}: 6.\ No medical advice
  \end{itemize}
\end{tcolorbox}

\section{Fine-tuning Details}\label{app:lora_hyperparams}

For community-based and norm-violation-based SLMs, we perform Low-Rank Adaptation (LoRA) fine-tuning~\cite{hu2021lora} for 1 epoch on balanced samples from the positive and negative class labels. We use rank $r = 16$, $\alpha = 32$. We do not use any dropout. Further, we use a learning rate $lr=2e-4$ with a linear schedule and $5$ warmup steps, and the AdamW~\cite{loshchilov2017decoupled} optimizer with a weight decay of $0.01$.

For the Global SLM baselines, due to the significantly larger amount of data these models are fine-tuned on, we perform $2$ epochs of fine-tuning, keeping all other hyperparameters intact. We now outline our fine-tuning prompts for each scenario, inspired by \citet{zhan-etal-2025-slm}.

\begin{tcolorbox}[
  title=\textbf{Community Expert SLM Prompt},
  colback=gray!10,
  colframe=gray!60!black,
  fonttitle=\bfseries,
  fontupper=\ttfamily\small, 
  rounded corners,
  breakable 
]
You are acting as a moderator for the r/\{SUBREDDIT\} subreddit. You will be given a comment from Reddit and the rules deemed suitable to arrive at a moderation outcome, and your task is to determine if the given text is undesirable or not based on the information provided to you.\\

\medskip
Here is a comment from a Reddit conversation thread, the context (preceding comment), and the associated subreddit rules.\\

\medskip
\textbf{Context:} \\
\{CONTEXT\}

\medskip
\textbf{Comment:} \\
\{COMMENT\}

\medskip
\textbf{Rules:} \\
\{RULES\}

\medskip
Determine whether the provided text is undesirable or not. Answer with `True' or `False'.\\

\medskip
\#\#\# Your Response:
\end{tcolorbox}

\begin{tcolorbox}[
  title=\textbf{Norm-Violation Expert SLM Prompt},
  colback=gray!10,
  colframe=gray!60!black,
  fonttitle=\bfseries,
  fontupper=\ttfamily\small,
  rounded corners,
  breakable
]
You are acting as a moderator for the r/\{SUBREDDIT\} subreddit. You will be given a comment from Reddit and the community norm deemed suitable to arrive at a moderation outcome, and your task is to determine if the given text violates the provided norm or not based on the information provided to you.\\

\medskip
Here is a comment from a Reddit conversation thread, the context (preceding comment), and the associated community norm.\\

\medskip
\textbf{Context:} \\
\{CONTEXT\}

\medskip
\textbf{Comment:} \\
\{COMMENT\}

\medskip
\textbf{Norm:} \\
\{NORM\}

\medskip
Determine whether the provided text is undesirable or not. Answer with `True' or `False'.\\

\medskip
\#\#\# Your Response:
\end{tcolorbox}

Finally, for the RoBERTa-base models fine-tuned for classification as part of the \allocate{} operators, we use fine-tune the model for 3 epochs with a learning rate $lr = 1e-5$ and maximum sequence length of $512$ tokens.

\section{Explanation Prompt Design}\label{app:explanation}

In this section, we report our prompt used for generating \momo{} explanations:

\begin{tcolorbox}[
  title=\textbf{\momonv{} Explain Prompt},
  colback=green!10,
  colframe=green!60!black,
  fonttitle=\bfseries,
  fontupper=\sffamily\small,
  rounded corners,
  breakable
]
\#\#\# System:\\
You are ``MoMoE-Explain'', an assistant that writes short, moderator‑facing rationales for AI-based content moderation decisions.\\
- Audience: Experienced Reddit moderators.\\
- Style: concise, neutral, no technical jargon, no private model thoughts.\\
- Output JSON keys in this exact order: Summary, Key Points, Trace.\\

\medskip
\#\#\# User:\\
Here is the decision trace for a comment: \{TRACE\}\\

\medskip
Generate:\\
1. \textbf{Summary}: $\le25$ words stating outcome recommendation ('Remove', 'Review', 'Keep'), Key norm violated, Consensus-level among experts ('Low', 'High').  

\medskip
2. \textbf{Key Points:} 2 bullet points ($\le10$ words each) covering: \\ 
- Top expert: <Name\_of\_Expert> (<Weight>)\\
- <Level\_of\_Consensus> consensus: X/5 experts - <Recommendation>

\medskip
3. \textbf{Trace:}\\
- Decision: ``<Decision>''\\
- Confidence: <\momo{} confidence\_score>\\
- Salient Spans: [``<span\_1>'', ``<span\_2'']

\medskip
For `Salient Spans', identify upto three specific sequence within the comment that likely led to the moderation outcome, keeping in mind the top experts that voted for its removal. If the outcome is to `Keep' the comment, leave the Salient Span list empty.\\

Respond only with valid JSON.
\end{tcolorbox}

We provide the model with three-shot exemplars of a TRACE and generated explanations, covering all decision cases in order to provide the model with the precise format expected.

\begin{table*}[t]
  \centering
  \small
  \sffamily
  \setlength{\tabcolsep}{6pt}
\resizebox{\linewidth}{!}{
  \begin{tabular}{>{\centering\arraybackslash}p{0.9cm}l
                  c
                  c c
                  c c
                  c c
                  c c}
    \toprule
    \rowcolor{blue!10}
    & \textbf{Subreddit} & \textbf{Imb.\ (\%)} &
    \textbf{Llama AUC} & \textbf{Llama F1} &
    \textbf{Mistral AUC} & \textbf{Mistral F1} &
    \textbf{G‐Llama AUC} & \textbf{G‐Llama F1} &
    \textbf{G‐Mistral AUC} & \textbf{G‐Mistral F1} \\
    \midrule
    \multirow{4}{*}{\rotatebox[origin=c]{90}{\textbf{Maj. Vote}}}
      & r/Overwatch       & 5  & 0.87 $\pm$ 0.02 & 0.46 $\pm$ 0.01 & \textbf{0.90 $\pm$ 0.02} & \textbf{0.49 $\pm$ 0.01} & 0.82 $\pm$ 0.02 & 0.44 $\pm$ 0.01 & 0.85 $\pm$ 0.02 & 0.47 $\pm$ 0.01 \\
      & r/hillaryclinton  & 5  & 0.60 $\pm$ 0.06 & 0.44 $\pm$ 0.02 & \textbf{0.71 $\pm$ 0.04} & \textbf{0.47 $\pm$ 0.02} & 0.55 $\pm$ 0.05 & 0.41 $\pm$ 0.02 & 0.66 $\pm$ 0.04 & 0.45 $\pm$ 0.02 \\
      & r/Overwatch       & 10 & 0.86 $\pm$ 0.01 & 0.52 $\pm$ 0.01 & \textbf{0.89 $\pm$ 0.02} & \textbf{0.56 $\pm$ 0.01} & 0.81 $\pm$ 0.02 & 0.50 $\pm$ 0.01 & 0.84 $\pm$ 0.02 & 0.54 $\pm$ 0.01 \\
      & r/hillaryclinton  & 10 & 0.60 $\pm$ 0.03 & 0.47 $\pm$ 0.01 & \textbf{0.70 $\pm$ 0.02} & \textbf{0.51 $\pm$ 0.01} & 0.55 $\pm$ 0.03 & 0.44 $\pm$ 0.01 & 0.65 $\pm$ 0.02 & 0.48 $\pm$ 0.01 \\
    \midrule
    \multirow{4}{*}{\rotatebox[origin=c]{90}{\textbf{Dot Prod.}}}
      & r/Overwatch       & 5  & 0.74 $\pm$ 0.01 & 0.43 $\pm$ 0.01 & \textbf{0.76 $\pm$ 0.01} & \textbf{0.44 $\pm$ 0.01} & 0.69 $\pm$ 0.01 & 0.40 $\pm$ 0.01 & 0.71 $\pm$ 0.01 & 0.42 $\pm$ 0.01 \\
      & r/hillaryclinton  & 5  & 0.57 $\pm$ 0.05 & 0.34 $\pm$ 0.01 & \textbf{0.63 $\pm$ 0.02} & \textbf{0.31 $\pm$ 0.01} & 0.52 $\pm$ 0.04 & 0.31 $\pm$ 0.02 & 0.58 $\pm$ 0.03 & 0.28 $\pm$ 0.02 \\
      & r/Overwatch       & 10 & 0.73 $\pm$ 0.01 & 0.49 $\pm$ 0.01 & \textbf{0.75 $\pm$ 0.01} & \textbf{0.51 $\pm$ 0.01} & 0.68 $\pm$ 0.02 & 0.46 $\pm$ 0.02 & 0.70 $\pm$ 0.01 & 0.48 $\pm$ 0.02 \\
      & r/hillaryclinton  & 10 & 0.58 $\pm$ 0.02 & 0.38 $\pm$ 0.01 & \textbf{0.63 $\pm$ 0.02} & \textbf{0.36 $\pm$ 0.01} & 0.53 $\pm$ 0.02 & 0.35 $\pm$ 0.01 & 0.58 $\pm$ 0.01 & 0.33 $\pm$ 0.01 \\
    \bottomrule
  \end{tabular}}
  \caption{Performance in terms of AUC and Macro-F1 of Llama‐ and Mistral‐based \momoco{} models under majority‐vote and dot‐product aggregation on class‐imbalanced test splits (5 \% and 10 \% “removed” labels). “Global SLMs’’ denote the strongest single‐model baselines, fine-tuned on $\mathcal{D}_\text{Community}$. Values are \texttt{mean} $\pm$ sd over 10 runs.}
  \label{tab:majority_dotproduct_results}
\end{table*}

\section{Further Discussion on MoMoE Performance Across Target Subreddits}\label{app:mistral-spread}

In this section, we provide a deeper discussion of the performance and differences between Mistral-based \momoco{} and \momonv{} on target communities, shown in \autoref{fig:momoe_f1_scores}.  All significance levels are from Welch's $t-$test~\cite{welch1947generalization}.

The Mistral-based \momoco{}, similar to the case of Llama, shows a much wider range of performance, with F1 scores ranging from $0.52$ for \textit{r/hillaryclinton} with the classification allocation strategy to as high $0.81$ for \textit{r/Overwatch} with the similarity allocation strategy. Overall, \momoco{} with similarity allocation achieves a mean F1 score of $0.71~(\pm0.06)$, while with classification allocation it achieves $0.67~(\pm0.07)$ ($p<0.05$). \momonv{} again shows contrasting, consistent performance across all subreddits, with F1 scores from $0.66$ for \textit{r/gonewild} with classification-based allocation to $0.72$ for \textit{r/PoliticalDiscussion} with classification-based allocation. Both allocation strategies for \momonv{} yield similar overall performance with mean F1 scores of $0.67~(\pm0.00)$ for similarity and $0.67~(\pm0.01)$ for classification-based allocation ($n.s.$). 

\section{Precision-Recall Trade-offs}\label{app:pr-tradeoff}

\begin{figure}[t]
    \centering
    \includegraphics[width=\linewidth]{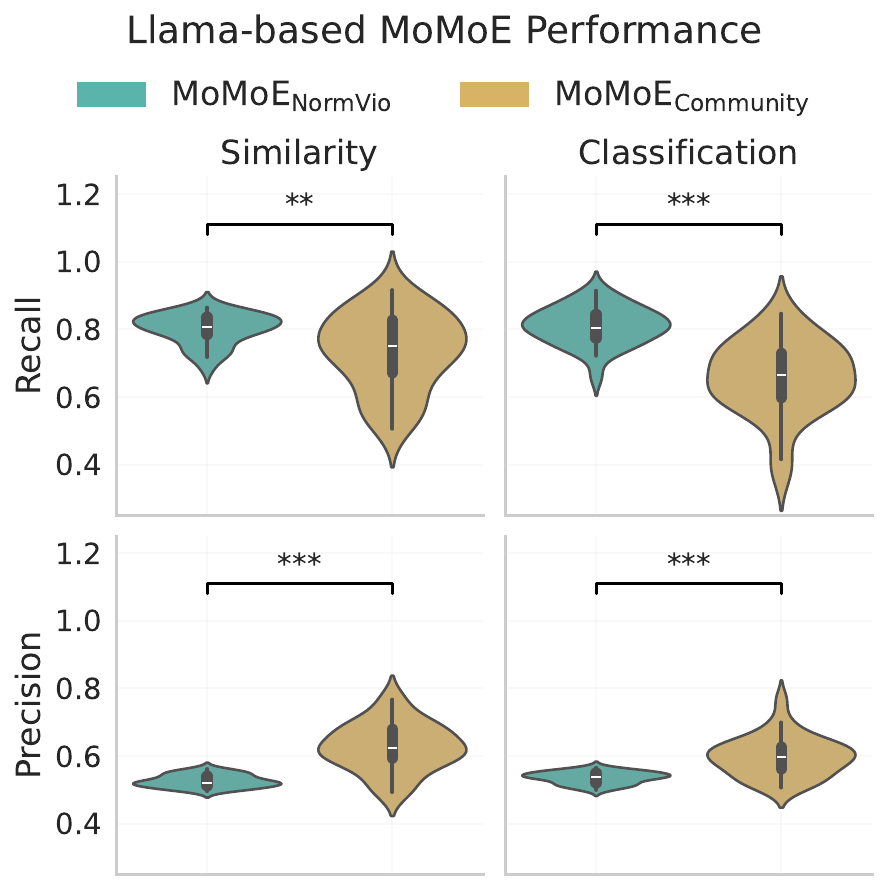}
    \caption{Comparison of precision-recall trade-offs with Llama-based \momoco{} with  \momonv{} using a dot-product aggregation. We observe that \momonv{} has higher recall compared to \momoco{} (mean difference $\approx0.06$), whereas in terms of precision, \momoco{} outperforms \momonv{} (mean difference $\approx0.08$). \small(${}^\ast$$p<0.05$, ${}^{\ast\ast}$$p<0.01$, ${}^{\ast\ast\ast}$$p<0.001$)}
    \label{fig:momoe_pr_tradeoff}
    \vspace{-16pt}
\end{figure}

We saw that both \momoco{} and \momonv{} perform competitively in comparison to baselines, while \momoco{} generally outperforms \momonv{}. We discuss here two kinds of precision-recall trade-offs with \momo{} in \autoref{fig:momoe_pr_tradeoff}.

First, we observe that the recall of both \momoco{} and \momonv{} is higher than their precision, which in combination with our existing results highlights that \momo{} may be overaggressive, flagging potentially violating comments rather than erring on the side of caution. In contrast, the precision and recall of LLMs on the target communities was $0.78$ and $0.44$ for GPT-4o-mini, and $0.82$ and $0.38$ for GPT-4o, respectively. This observation is in line with that of \citet{zhan-etal-2025-slm}, who found that SLMs prioritize potentially harmful content even at the cost of over-flagging.

Second, within the two types of ensembles \momoco{} and \momonv{} we observe that on recall, \momonv{} outperforms \momoco{}, with a recall of $0.80 (\pm 0.05)$ compared to that of \momoco{} at $0.73(\pm0.11)$ for similarity-based allocation, and $0.81(\pm0.06)$ in comparison to $0.66(\pm0.11)$ for \momoco{}. With precision on the other hand, we observe that \momoco{} outperforms \momonv{}, with a precision of $0.63 (\pm 0.07)$ compared to that of \momonv{} at $0.53(\pm0.02)$ for similarity-based allocation, and $0.60(\pm0.06)$ in comparison to $0.54(\pm0.02)$ for \momonv{}. All differences are significant by Welch's $t-$test. This highlights that although both ensembles are over-aggressive at removing comments, this tendency is particularly enhanced in \momonv{}.

For practitioners, this means that the \predict{} component \momonv{} in a standalone manner is more suited for a comment triaging scenario where a human moderator will oversee decisions, rather than automated moderation settings. This would ensure that community members are not wrongfully punished with their benign comments being removed. We see the same trend with Mistral-based \momo{} as well.

\section{Performance on Imbalanced Test Split}\label{app:imbalanced-results}

In this section we report the performance of the best performing \predict{} configuration, \momoco{} on imbalanced test split on the worst (\textit{r/hillaryclinton}) and best performing (\textit{r/Overwatch}) subreddits. \autoref{tab:majority_dotproduct_results} shows performance of Llama- and Mistral-based \momoco{} compared to the best performing baseline of G-Llama$_\text{Community}$ and G-Mistral$_\text{Community}$ in terms of AUC and Macro-F1 scores. Prior work has shown that the proportion of comments that actually violate community norms in the real world are around 6-7\% of all comments~\cite{park2022measuring}. We therefore test on two imbalance thresholds of 5\% and 10\% violating comments, and the remaining non-violating comments over $10$ random seeds. Since both classification and similarity-based allocation performed very similarly in the case of \momoco{}, we report here the results for classification-based allocation.

Similar to the case of balanced test split, we observe that Mistral-based \momoco{} performs the best in terms of both AUC and F1 scores, followed by Llama-based \momoco{}, both of which outperform the Global SLM baselines. We also again observe that majority vote based aggregation works better than dot product aggregation. These results indicate that \momo{} continues to show superior performance on target communities even under a more realistic imbalanced data scenario.

\section{Compute Resources}

All experiments on open-source models were run on internal GPU servers equipped with 4xNVIDIA A100 and 3xNVIDIA A40. The experiments with the OpenAI models cost about 100 USD.

\end{document}